%% file: top.tex
\documentclass[runningheads]{llncs}
\usepackage{graphicx}

\usepackage{comment}
\usepackage[utf8]{inputenc}
\usepackage{epsfig}
\usepackage{breqn}
\usepackage{amssymb}
\usepackage{multirow}
\usepackage{hyperref}
\usepackage{booktabs}
\usepackage{hhline}
\usepackage{tikz}
\usepackage{pgfplots}
\pgfplotsset{compat=1.14}

\input{macro.tex}

\begin{document}
\pagestyle{headings}
\mainmatter
\def\ECCVSubNumber{4918}  %

\title{Learned Vertex Descent: \newline A New Direction for 3D Human Model Fitting}

\titlerunning{Learned Vertex Descent: A New Direction for 3D Human Model Fitting}

\author{Enric Corona$^{1}$ \and \hspace{0.5mm}
Gerard Pons-Moll$^{2,3}$ \and \\ \hspace{0.5mm} Guillem Aleny\`a$^{1}$ \and Francesc Moreno-Noguer$^{1}$}
\institute{\hspace{-4mm}${}^{1}$Institut de Robòtica i Informàtica Industrial, CSIC-UPC, Barcelona, Spain\\${}^{2}$University of T\"{u}bingen, Germany, ${}^{3}$Max Planck Institute for Informatics, Germany}

\authorrunning{Corona et al.}

\maketitle

\input{section_abstract}

\input{section_introduction}

\input{section_related_work}

\input{section_method}

\input{section_experiments}

\input{section_conclusion}

{\small
\paragraph{Acknowledgements}
This work is supported in part by the
Spanish government with the project MoHuCo PID2020-120049RB-I00, the Deutsche Forschungsgemeinschaft (DFG, German Research Foundation) - 409792180 (Emmy Noether Programme,
project: Real Virtual Humans) and German Federal Ministry of Education and Research (BMBF): Tübingen AI Center, FKZ: 01IS18039A. Gerard Pons-Moll is a member of the Machine Learning Cluster of Excellence, EXC number 2064/1 – Project number 390727645. We thank
NVIDIA for donating GPUs.
}

\bibliographystyle{splncs04}
\bibliography{references}

\input{supplementary}

\end{document}

%% file: macro.tex
\newcommand{\Modelname}{LVD}

\def\eg{\emph{e.g. }}
\def\etal{\emph{et al}}

\newcommand{\mat}[1]{\mathbf{#1}}

\newcommand{\real}[0]{\mathbb{R}}

\newcommand{\be}{\mathbf{e}}
\newcommand{\bff}{\mathbf{f}}

\newcommand{\bp}{\mathbf{p}}

\newcommand{\bv}{\mathbf{v}}

\newcommand{\bx}{\mathbf{x}}

\newcommand{\bF}{\mathbf{F}}

\newcommand{\bI}{\mathbf{I}}

\newcommand{\bV}{\mathbf{V}}

\newcommand{\bX}{\mathbf{X}}

\newcommand{\bhv}{\hat{\mathbf{v}}}
\newcommand{\bhV}{\hat{\mathbf{V}}}

%% file: section_abstract.tex
\begin{abstract}
We propose a novel optimization-based paradigm for 3D human model fitting on images and scans. In contrast to existing  approaches that directly regress the parameters of a low-dimensional statistical body model (\eg SMPL) from input images, we train an ensemble of per vertex neural fields network. The network predicts, in a distributed manner, the vertex descent direction towards the ground truth, based on neural features extracted at the current vertex projection.   
At inference, we employ this network, dubbed \Modelname, within a gradient-descent optimization pipeline until its convergence, which typically occurs in a fraction of a second even when initializing all vertices into a single point.  An exhaustive evaluation demonstrates that our approach is able to capture the underlying body  of clothed people with very different body shapes, achieving a significant improvement compared to state-of-the-art. LVD is also applicable to 3D model fitting of humans and hands, for which we show a significant improvement to the SOTA with a much simpler and faster method. Code is released at \url{https://www.iri.upc.edu/people/ecorona/lvd/}

\end{abstract}

%% file: section_introduction.tex
\section{Introduction}

Fitting 3D human models to data (single images / video / scans) is a highly ambiguous problem. The standard approach to overcome this is by introducing  statistical shape priors ~\cite{anguelov2005scape,smpl,xu2020ghum} controlled by a reduced number of parameters. Shape recovery then entails at estimating these parameters from data. There exist two main paradigms for doing so.

On the one side, optimization-based methods iteratively search for the model parameters that best match available image cues, like 2D keypoints~\cite{smplx,bogo2016keep,arnab2019exploiting,spin},  silhouettes~\cite{lassner2017unite,sigal2007combined} or dense correspondences~\cite{guler2019holopose}. On the other side, data-driven regression methods for mesh recovery leverage deep neural networks to directly predict the model parameters from the input~\cite{hmr,guler2019holopose,texturepose,expose,georgakis2020hierarchical,alldieck2019tex2shape,omran2018neural}. In between these two streams, there are recent approaches that build hybrid methods combining optimization-regression schemes~\cite{zanfir2021neural,joo2020exemplar,spin,song2020human}. 

\input{figure_teaser}

Regardless of the inference method, optimization or regression, and input modality, 2D evidence based on the entire image, keypoints, silhouettes, pointclouds, all these previous methods aim at estimating the parameters of a low-dimensional model (typically based on SMPL~\cite{smpl}). %
However, as we will show in the experimental section, these models
struggle in capturing detailed body shape, specially for morphotypes  departing from the mean (overweight or skinny people) or when the person is wearing loose clothing. 
We hypothesize that this is  produced by two main reasons: 1) the  models induce a bias towards the mean shape; and 2) the mapping from local image / pointcloud features to \emph{global shape} parameters is highly non-linear. This makes optimization-based approaches prone to get stuck at local minima and have slow run times. Global shape regression methods lack the error-feedback loop of optimization methods (comparing the current estimate against image / scan input), and hence exhibit an even more pronounced bias towards mean shapes. Overcoming this problem would require immense amounts of training data, which is infeasible for 3D bodies. 

To recover more detail, recent works regress or optimize a set of displacements on top of SMPL global shape~\cite{alldieck2018video,alldieck19cvpr,alldieck2019tex2shape,bhatnagar2019multi,patel20tailornet}, local surface elements~\cite{scale} or points~\cite{ma2021power}.
Like us, ~\cite{kolotouros2019convolutional} by-pass the regression of global shape parameters and regress model vertices direclty. However, 
similar to displacement-based methods~\cite{alldieck19cvpr,alldieck2019tex2shape}, the proposed regression scheme~\cite{kolotouros2019convolutional} predicts the position of all points in one single pass and lacks an error-feedback loop. Hence, these methods regress a global shape in one pass based on global image features and also suffer from bias towards the mean. Works based on implicit surfaces~\cite{chibane2020neural,saito2019pifu,saito2020pifuhd} address these limitations by making point-wise distributed predictions. Being more local, they require less training data. However, these methods do not produce surfaces with a coherent parameterization (e.g. SMPL vertices), and hence control is only possible with subsequent model fitting, which is hard if correspondences are not known~\cite{ipnet,loopreg,huang2020arch,he2021arch++}.

In this paper, we propose a significantly different approach to all prior model fitting methods.
Inspired by classical model-based fitting, where image gradients drive the direction of vertices and in turn global shape parameters, we propose to iteratively learn where 3D vertices should move based on neural features.
For that purpose, we devise a novel data-driven optimization in which an ensemble of per-vertex neural fields is trained to predict the optimal 3D vertex displacement towards the ground-truth, based on local neural features extracted at the current vertex location. 
We dub this network LVD, from `Learned Vertex Descent'. At inference, given an input image or scan, we initialize all mesh vertices into a single point and iteratively query LVD to estimate the vertex displacement in a gradient descent manner.

We conduct a thorough evaluation of the proposed learning-based optimization approach. 
Our experiments reveal that LVD combines the advantages of classical optimization and learning-based methods. 
LVD captures off-mean shapes significantly more accurately than all prior work, unlike optimization approaches it does not suffer from local minima, and converges in just 6 iterations. We attribute the better performance to the \emph{distributed per-vertex predictions} and to the \emph{error feedback loop} -- the current vertex estimate is iteratively verified against the image evidence, a feature present in all optimization schemes but missing in learning-based methods for human shape estimation. 

We demonstrate the usefulness of LVD for the tasks of 3D human shape estimation from images, and 3D scan registration(see Fig~\ref{fig:teaser}). In both problems, we surpass existing approaches by a considerable margin.

\vspace{1mm}
\noindent Our key contributions can  be summarized as follows:
\begin{itemize}
\setlength{\itemsep}{0pt}

\item A novel learning-based optimization where vertices \emph{descent} towards the correct solution according to learned neural field predictions. This optimization is fast, does not require gradients and hand-crafted objective functions, and is not sensitive to initialization.
\item We empirically show that our approach achieves state-of-the-art results in the task of human shape recovery from a single image.
\item The LVD formulation can be readily adapted to the problem of 3D scan fitting. We also demonstrate state-of-the-art results on fitting 3D scans of full bodies and hands. 
\item By analysing the variance of the learned vertex gradient in local neighborhoods we can  extract uncertainty information about the reconstructed shape. This might be useful for subsequent downstream applications that require confidence measures on the estimated body shape.
\end{itemize}

%% file: figure_teaser.tex
\begin{figure*}
\vspace{-1mm}
\includegraphics[width=1\linewidth, trim={0.1cm, 0.1cm 0.1cm 0.1cm}, clip=true]{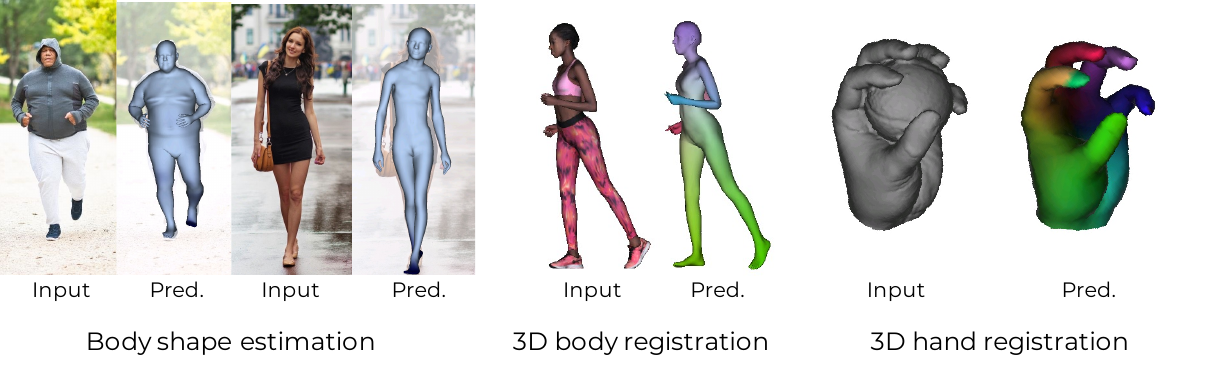}\\
\vspace{-7mm}
\caption{
\small{{\bf Learned Vertex Descent (LVD)} is  a novel optimization strategy in which a network leverages local image or volumetric features to iteratively predict per-vertex directions towards an optimal body/hand surface.
The proposed approach is directly applicable to different tasks with minimal changes on the network, and we show it can fit a much larger variability of  body shapes than previous state-of-the-art. The figure depicts results on the three tasks where we have evaluated LVD: body shape reconstruction from a single image, and 3D fitting of body and hand scans.%
\label{fig:teaser}
}}
\vspace{-5mm}
\end{figure*}

%% file: section_related_work.tex
\section{Related work}

\subsection{Parametric models for 3D body reconstruction }
The {\em de-facto} approach for reconstructing human shape and pose is by estimating the parameters of a low-rank generative model~\cite{smpl,smplx,xu2020ghum,mano}, being SMPL~\cite{smpl} or SMPL-X~\cite{smplx} the most well known. We next describe the approaches to perform model fitting from images.

\vspace{1mm}
\noindent{\bf Optimization.}
Early approaches on human pose and shape estimation from images 
used optimization-based approaches to estimate the model parameters from 2D image evidence. Sigal~\etal~\cite{sigal2007combined} did so for the SCAPE~\cite{anguelov2005scape} human model, and assuming 2D input silhouettes. Guan~\etal~\cite{guan2009estimating}, combined  silhouettes with manually annotated 2D skeletons. 
More recently, the standard optimization relies on 2D   skeletons~\cite{bogo2016keep,smplx,song2020human}, estimated by off-the-shelf and robust deep methods~\cite{openpose}. This is typically accompanied by additional pose priors to ensure anthropomorphism of the retrieved pose~\cite{bogo2016keep,smplx}. %
Subsequent works have devised approaches to obtain better initialization from image cues~\cite{spin}, more efficient optimization pipelines~\cite{song2020human}, focused on multiple people~\cite{dong2021shape} or extended the approach to multi-view scenarios~\cite{li20213d,dong2021shape}.

While optimization-based approaches do not require images with 3D annotation for training and achieve relatively good registration of details to 2D observations, they tend to suffer from the non-convexity of the problem, being slow and falling into local minima unless provided with a good initialization and accurate 2D observations.
In this work, we overcome both these limitations. From one side we only use as input a very coarse person segmentation and image features obtained with standard encoder-decoder architectures. And from the other side, the learned vertex displacements help the optimizer to converge to good solutions (empirically observed) in just a few iterations. On the downside, our approach requires 3D training data, but as we will show in the experimental section, by using synthetic data we manage to generalize well to real images.

\vspace{1mm}
\noindent{\bf Regression.} Most current approaches on human body shape recovery consider the direct regression of the shape and pose parameters of the SMPL model~\cite{guler2019holopose,arnab2019exploiting,vibe,kolotouros2019convolutional,texturepose,georgakis2020hierarchical,prohmr,frankmocap,sengupta2020synthetic,sengupta2021hierarchical,choutas2022accurate}. As in  optimization-based methods, different sorts of 2D image evidence have been used, \eg keypoints~\cite{lassner2017unite}, keypoints plus silhouette~\cite{PavlakosCVPR2018} or part segmentation maps~\cite{omran2018neural}. More recently, SMPL parameters have been regressed directly from entire images encoded by pre-trained deep networks (typically ResNet-like)~\cite{hmr,guler2019holopose,texturepose,expose,georgakis2020hierarchical}. 
However, regressing the parameters of a low-dimensional parametric model from a single view is always a highly ambiguous problem.
This is alleviated by recent works that explore the idea of using hybrid approaches combining optimization and regression~\cite {zanfir2021neural,joo2020exemplar,spin,song2020human,moon2020i2l,lin2021end}. Very recently,~\cite{prohmr} proposed regressing a distribution of parameters instead of having a regression into a single pose and shape representation.
In any event, all these works still rely on representing the body shape through low-rank models. 

We argue that other shape representations are necessary to model body shape details. This was already discussed in~\cite{kolotouros2019convolutional}, which suggested representing the body shape using all vertices of a template mesh. We will follow the same spirit, although in a completely different learning paradigm. Specifically~\cite{kolotouros2019convolutional} proposed  regressing all points of the body mesh in one single regression pass of a Graph Convolutional Network. This led to noisy outputs that required from post-processing step to smooth the results by fitting the SMPL model. Instead, we propose a novel optimization framework, that leverages on a pre-learned prior that maps image evidence to vertex displacements towards the body shape. We will show that despite its simplicity, this approach surpasses by considerable margins all prior work, and provides smooth while accurate meshes without any post-processing.

\subsection{Fitting scans}

Classical ICP-based has been used for fitting SMPL with no direct correspondences~\cite{pishchulin2017building,faust,dynamic_faust,dyna,coregistration,nonrigid_registrations} or for registration of garments~\cite{ponsmollSIGGRAPH17clothcap,bhatnagar2019multi,deepwrinkles}. Integrating additional knowledge such as pre-computed 3D joints, facial key points~\cite{alldieck2019learning} and body part segmentation~\cite{ipnet} significantly improves the registration quality but these pre-processing steps are prone to error and often require human supervision. Other works initialize correspondences with a learned regressor ~\cite{taylor2012vitruvian,pons2015metric,groueix20183d} and later optimize model parameters. Like us, more recent methods also propose predicting correspondences~\cite{loopreg} or body part labels~\cite{ipnet} extracted via learnt features. Even though we do not explicitly propose a 3D registration method, LVD is a general algorithm that predicts parametric models. By optimizing these predictions without further correspondences, we surpass other methods that are explicitly designed for 3D registration.

\subsection{Neural fields for parametric models}

Neural fields~\cite{smplicit,prokudin2021smplpix,pan2019deep,corona2022lisa,zhou2022toch,tiwari22posendf} have recently shown impressive results in modeling 3D human shape~\cite{snarf:chen:iccv21,zheng2021pamir,deprelle2019learning,pan2019deep,nasa,niemeyer2019occupancy}. However, despite providing the level of detail that parametric models do not have,  they are  computationally expensive and difficult to integrate within pose-driven applications given the lack of correspondences.  Recent works have already explored possible integrations between implicit and parametric representations for the tasks of 3D reconstruction~\cite{huang2020arch,xie22chore}, clothed human modeling~\cite{scanimate,scale,ma2020learning}, or human rendering~\cite{neuralbody}.

We will build upon this  direction by framing our method in the pipeline of neural fields. Concretely, we will take the vertices of an unfit mesh and use image features to learn their optimal displacement towards the optimal body shape.

%% file: section_method.tex
\section{Method}\label{sec:approach}
We next present our new paradigm for fitting 3D human models. For clarity, we will describe  our approach   in the problem of 3D human shape reconstruction from a single image. Yet, the  formulation we present here is generalizable to the problem of fitting 3D scans, as we shall demonstrate in the experimental section.

\vspace{-1mm}
\subsection{Problem formulation}
Given a single-view image $\bI\in\real^{H\times W}$ of a person our goal is to reconstruct his/her full body. We represent the body using a 3D mesh $\bV\in\real^{N\times 3} $ with $N$ vertices. For convenience (and compatibility with SMPL-based downstream algorithms) the mesh topology will correspond to that of the SMPL model, with $N=6.890$ vertices and triangular connectivity ($13.776$ faces). 
It is important to note that our method operates on the vertices directly and hence it is applicable to other models (such as hands~\cite{mano}). In particular, we do not use the low dimensional pose and shape parameterizations of such models.

\input{figure_arch}

\subsection{LVD: Learning Vertex Descent}
We solve the model fitting problem via an iterative optimization approach with learned vertex descent. Concretely, let $\bv_i^t$ be the $i$-th vertex of the estimated mesh $\bV$ at iteration $t$. 
Let us also denote by $\bF \in\real^{H'\times W'\times F}$ the pixel aligned image features, and by $\bff_i$ the $F$-dimensional vector of the specific features extracted at the projection of $\bv_i^t$ on the image plane.

We  learn a function $g(\cdot)$ that given the current 3D vertex position, and the image features at its 2D projection, predicts the magnitude and direction of steepest descent towards the ground truth location of the $i$-th vertex, which we shall denote as $\bhv_i$. Formally:
\begin{equation}
g: (\bv_i^t,\bff_i) \mapsto \Delta \bv_i\,.\label{eq:model}
\end{equation}
where  $\Delta \bv_i \in \real^3$ is a vector with origin at $\bv_i^t$ and endpoint at the ground truth $\bhv_i$. In practice, during training, we will apply a component-wise clipping to the ground truth displacements with threshold $\lambda$.
This stabilizes convergence during the first training iterations.

We learn the vertex descent function $g(\cdot)$ using a joint ensemble of per-vertex neural field networks, which we describe in Sect.~\ref{sec:arquitecture}. Once this mapping  is learned, we can define the following update rule for our learned optimization:
\begin{equation}
\bv_i^{t+1}= \bv_i^t+ \Delta \bv_i\;. \label{eq:iter}
\end{equation}
The reconstruction problem then entails iterating over Eq.~\ref{eq:iter}
until the convergence of $\Delta \bv_i $. Fig~\ref{fig:overview} depicts an overview of the approach. 

Note that in essence we are replacing the standard gradient descent rule  with a learned update that is locally computed  at every vertex.   As we will empirically demonstrate in the results section, despite its simplicity, the proposed approach  allows for fast and remarkable convergence rates, typically requiring only 4 to 6 iterations no matter how the mesh vertices are initialized.

\vspace{1mm}
\noindent{\bf Uncertainty estimation.} An interesting outcome of our approach is that it allows estimating the uncertainty of the estimated 3D shape, which could be useful  in downstream  applications that require a confidence measure.
For estimating the uncertainty of a vertex $\bv_i$, we compute the variance of the points after perturbing them and letting the network converge. After this process, we obtain the displacements $\Delta \bx_j^i$ between perturbed points $\bx_j$ and the mesh vertex $\bv_i$ predicted initially. We then define the uncertainty of $\bv_i$ as: 
\begin{equation}
    U(\bv_i)=\textrm{std}(\{\bx_j + \Delta \bx_{j}^{i}\}_{j=1}^M)\;.
\end{equation}
In Figs.~\ref{fig:teaser} and~\ref{fig:qualitative_results_image} we represent the uncertainty of the meshes in dark blue. Note that the most uncertain regions are typically localized on the feet and hands.

\subsection{Network architecture} \label{sec:arquitecture}

The LVD architecture has two main modules, one that is responsible of extracting local image features and the other of learning the optimal vertices' displacement. 

\vspace{1mm}
\noindent{\bf Local features.} Following recent approaches~\cite{saito2019pifu,saito2020pifuhd}, the local features $\bF$ are learned with an encoder-decoder Hourglass network trained from scratch. Given a vertex $\bv_i^t=(x_i^t,y_i^t,z_i^t)$ and the input image $\bI$, these features are estimated as:
\begin{equation}
f: (\bI,\pi(\bv_i^t),z_i^t) \mapsto \bff_i\;, \label{eq:feat}
\end{equation}
where  $\pi(\bv)$ is a weak perspective projection of $\bv$ onto the image plane. We condition $f(\cdot)$ with the depth $z_i^t$ of the vertex to generate depth-aware local features. A key component of LVD is Predicting vertex displacements based on local features, which have been shown to produce better geometric detail, even from small training sets~\cite{saito2019pifu,saito2020pifuhd,ifnet}. Indeed, this is one of our major differences compared to previous learning approaches for human shape estimation relying on parametric body models. These methods learn a mapping from a full image to global shape parameters (two disjoint spaces), which is hard to learn, and therefore they are unable to capture the local details. This results in poor image overlap between the recovered shape and the image as can be seen in Fig.~\ref{fig:teaser}.

\vspace{1mm}
\noindent{\bf Network field.} In order to implement the function $g(\cdot)$ in Eq.~\ref{eq:model} we follow recent neural field approaches~\cite{mescheder2019occupancy,deepsdf} and use a simple 3-layer MLP that takes as input the current estimate of each vertex $\bv_i^t$ plus its local $F$-dimensional local feature $\bff_i$  and predicts the displacement $\Delta \bv_i$.

\subsection{Training LVD}
Training the proposed model  entails learning the parameters of the functions  $f(\cdot)$ and $g(\cdot)$ described above. For this purpose, we will leverage a synthetic dataset of images of people under different clothing and body poses paired with  the corresponding SMPL 3D body registrations. We will describe this dataset in the experimental section. 

In order to train the network, we proceed as follows: Let us assume we are given a ground truth body mesh $\bhV=[\bhv_1,\ldots,\bhv_N]$ and its corresponding image $\bI$.  We then randomly sample $M$ 3D points  $\mathcal{X}=\{\bx_1,\ldots,\bx_M\}$, 
using a combination of points uniformly sampled in space and points distributed near the surface.
Each of these points, jointly with the input image $\bI$ is fed to the LVD model which predicts its displacement w.r.t. all ground truth SMPL vertices. Then, the loss associated with $\bx_i$ is computed as: 
\begin{equation}
\mathcal{L}(\bx_i)=\sum_{j=1}^N \|\Delta \bx_i^j - \hat{\Delta} \bx_i^j\|_1\;,
\end{equation}
where $\Delta \bx_i^j$ is the predicted displacement between $\bx_i$ and $\bhv_j$ and $\hat{\Delta} \bx_i^j$ the ground truth displacement. $\|\cdot\|_1$ is the L1 distance. Note that by doing this, we are teaching our network to predict the displacement of any point in space to all vertices of the mesh.  We found that this simple loss was sufficient to learn  smooth but accurate body prediction.  Remarkably, no additional regularization losses enforcing geometry consistency or anthropomorphism were required. 

The reader is referred to the Supplemental Material for additional implementation and training details.

\subsection{Application to 3D scan registration}
The pipeline we have just described can be readily applied to the problem of fitting the SMPL mesh to 3D scans of clothed people or fitting the MANO model~\cite{mano} to 3D scans of hands. The only difference will be in the feature extractor $f(\cdot)$ of Eq.~\ref{eq:feat}, which will have to account for volumetric features. That is, if $\bX$ is a 3D voxelized input scan, the feature extractor for a vertex $\bv_i$ will be defined as: 
\begin{equation}
f^{3D}: (\bX,\bv_i) \mapsto \bff_i\;, \label{eq:feat3D}
\end{equation}
where again, $\bff_i$ will be an $F$-dimensional feature vector. For the MANO model, the number of vertices of the mesh is $N=778$.
In the experimental section, we will show the adaptability of LVD to this scan registration problem.

\section{Connection to classical model based fitting}

Beyond its good performance, we find the connection of LVD to classical optimization based methods interesting, and understanding its relationship can be important for future improvements and extensions of LVD.    
Optimization methods for human shape recovery optimize model parameters to match image features such as correspondences~\cite{guan2009estimating,bogo2016keep,smplx}, silhouettes~\cite{sigal2007combined,sminchisescu2001covariance}. See~\cite{ponsmollModelBased} for an in-depth discussion of optimization-based model based fitting.

\input{figure_image_shape}

\vspace{1mm}
\noindent{\bf Optimization based.} These methods minimize a scalar error $e(\bp) \in \mathbb{R}$ with respect to human body parameters $\bp$. Such scalar error is commonly obtained from a sum of squares error $e = \be(\bp)^T\be(\bp)$. 
The error vector $\be \in \mathbb{R}^{dN}$ contains the $d$ dimensional residuals for the $N$ vertices of a mesh, which typically correspond to measuring how well the projected $i-th$ vertex in the mesh fits the image evidence (\emph{e.g}, matching color of the rendered mesh vs image color). 
To minimize $e$ one can use gradient descent, Gauss-Newton or Levenberg-Marquadt (LM) optimizer to find a descent direction for human parameters $\bp$, but ultimately the direction is obtained from local image gradients as we will show.  
Without loss of generality, we can look at the individual residual incurred by one vertex $\be_i \in \mathbb{R}^{d}$, although bear in mind that an optimization routine considers all residuals simultaneously (the final gradient will be the sum of individual residual gradients or step directions in the case of LM type optimizers). 
The gradient of a single residual can be computed as
\begin{equation}
  \nabla_{\bp} e_i = \frac{\partial (\be_i^T\be_i)}{\partial \bp} =  2 \left[\frac{\partial \be_i}{\partial \bv_i}\frac{\partial \bv_i}{\partial \bp}\right]^T \be_i 
\end{equation} 
where the matrices that play a critical role in finding a good direction
are the error itself $\be_i$, and $\frac{\partial \be_i}{\partial \bv_i}$ which is the Jacobian matrix of the $i$-th residual with respect to the $i$-th vertex
(the Jacobian of the vertex w.r.t. to parameters $\bp$ is computed from the body model and typically helps to restrict (small) vertex displacements to remain within the space of human shapes).
When residuals are based on pixel differences (common for rendering losses and silhouette terms) obtaining $\frac{\partial \be_i}{\partial \bv_i}$ requires computing image gradients via finite differences. Such classical gradient is only meaningful once we are close to the solution.

\input{figure_image_reconstruction}
\input{results_buff_quantitative}

\vspace{1mm}
\noindent{\bf Learned Vertex Descent.} In stark contrast, our neural fields compute a learned vertex direction, with image features that have a much higher receptive field than a classical gradient. 
This explains why our method converges much faster and more reliably than classical approaches. To continue this analogy, our network learns to minimize the following objective error (for a single vertex)
\begin{equation}
	e^{LVD}_i = \be_i^T \be_i = (\bv_i-\bv^{gt}_i)^T(\bv-\bv^{gt}_i)
\end{equation}
whose vertex gradient $\nabla_{\bv_i} e^{LVD}_i$ points directly to the ground truth vertex $\bv^{gt}_i$. In fact, our LVD is trained to learn the step size as well as the direction. What is even more remarkable, and surprising to us, is that we do not need a body model to constraint the vertices. That is, during optimization, we do not need to compute $\frac{\partial\bv_i}{\partial\bp}$, and project the directions to the space of valid human shapes. Since LVD has been learned from real human shapes, it automatically learns a prior, making the model very simple and fast during inference.

%% file: figure_arch.tex
\begin{figure*}[t!]
\vspace{-2mm}
\noindent\includegraphics[width=1\linewidth, trim={0.1cm, 0.1cm 0.1cm 0.1cm}, clip =true]{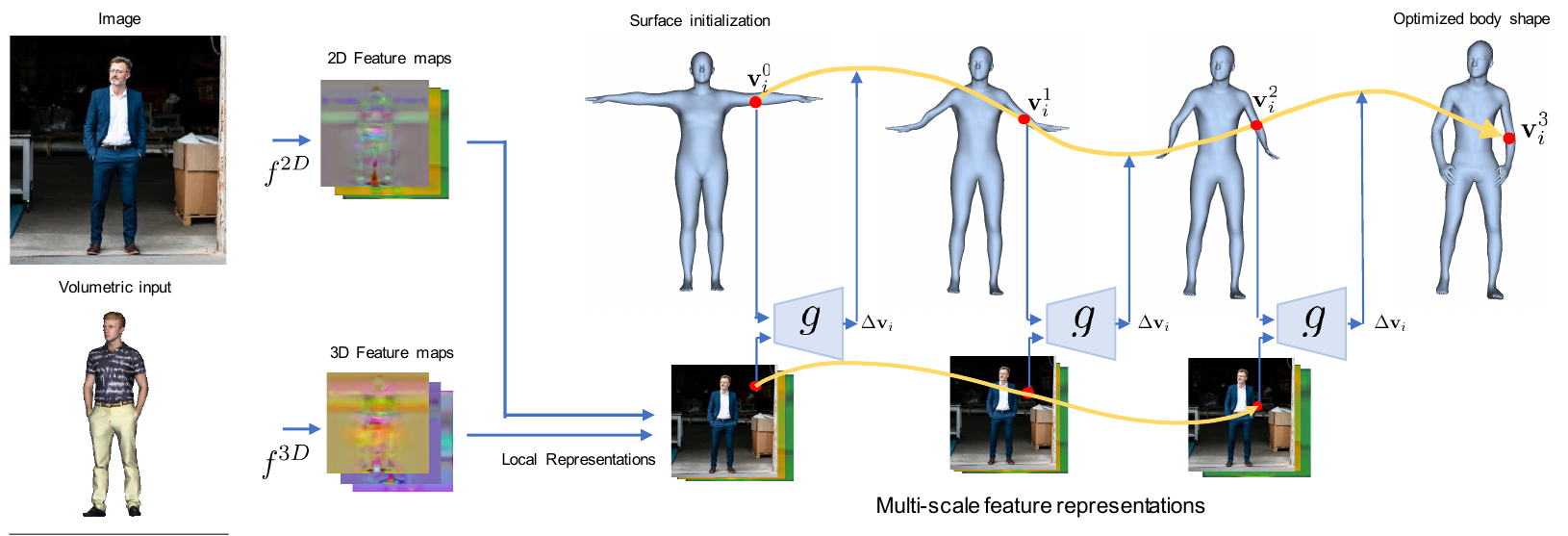}
\vspace{-6mm}
\caption{\small{
\Modelname~is a novel framework for estimation of 3D human body where local features drive the direction of vertices iteratively by predicting a per-vertex neural field. 
At each step $t$, $g$ takes an input vertex $\mat{v}_i^t$ with its corresponding local features, to predict the direction towards its groundtruth position. 
The surface initialization here follows a T-Posed body, but the proposed approach is very robust to initialization.
\label{fig:overview}
}}
\vspace{-3mm}
\end{figure*}

%% file: figure_image_shape.tex
\begin{figure}[t]
\vspace{-2mm}
\noindent\includegraphics[width=1\linewidth, trim={0.1cm, 0.1cm 0.1cm 0.1cm}, clip =true]{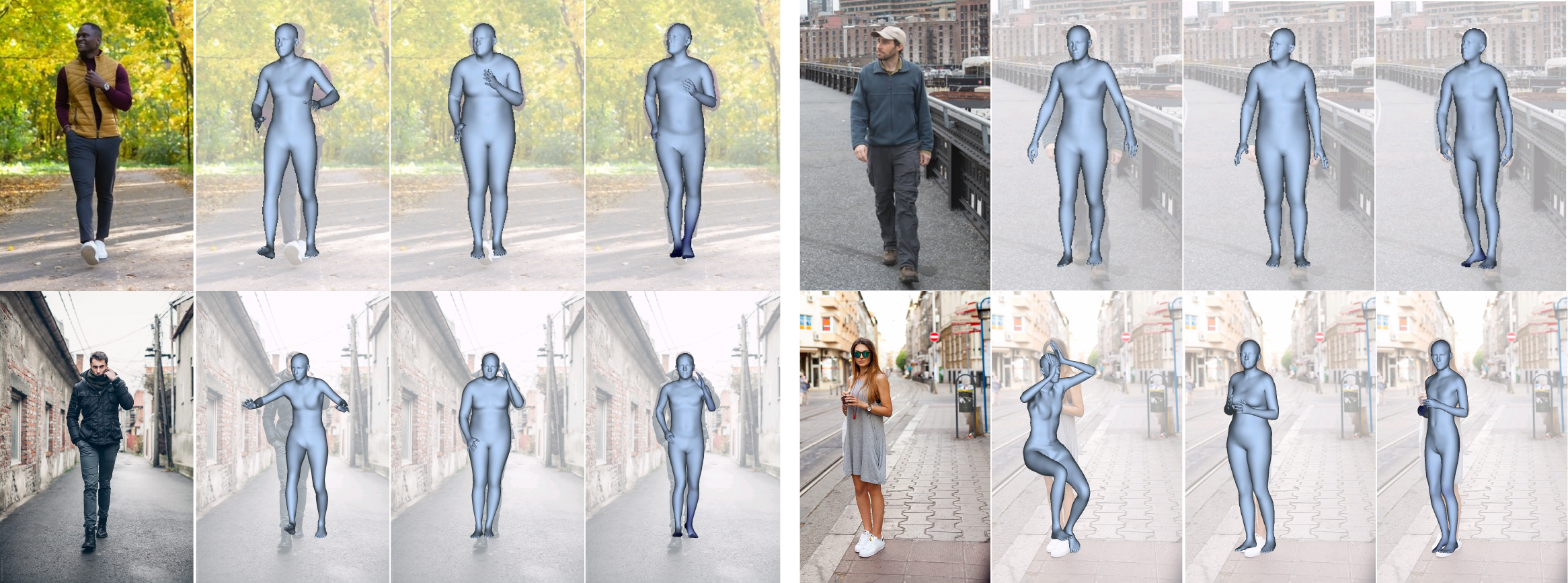}\\
\put(10, 0){\scriptsize{Input}}
\put(42, 0){\scriptsize{ResNet~\cite{resnet}}}
\put(86, 0){\scriptsize{Sengupta~\cite{sengupta2021hierarchical}}}
\put(145, 0){\scriptsize{LVD}}
\put(182, 0){\scriptsize{Input}}
\put(217, 0){\scriptsize{ResNet~\cite{resnet}}}
\put(260, 0){\scriptsize{Sengupta~\cite{sengupta2021hierarchical}}}
\put(319, 0){\scriptsize{LVD}}
\vspace{-2mm}
\caption{\small{
{\bf Comparison of LVD to body shape estimation baselines.} As a first baseline, we train a ResNet~\cite{resnet} (using the same data as for LVD) to predict the SMPL parameters. This approach fails to generalize to novel poses and shapes. We also compare LVD to Sengupta~\etal~\cite{sengupta2021hierarchical}, which perform well on real images, even though the predicted shapes do not fit perfectly the silhouettes of the people. See also quantitative results in Table~\ref{tab:quantitative_buff}.
\label{fig:qualitative_results_shape}
}}

\vspace{-5mm}
\end{figure}

%% file: figure_image_reconstruction.tex
\begin{figure}[t]
\centering
\includegraphics[width=1\linewidth, trim={0.1cm, 0.1cm 0.1cm 0.1cm}, clip =true]{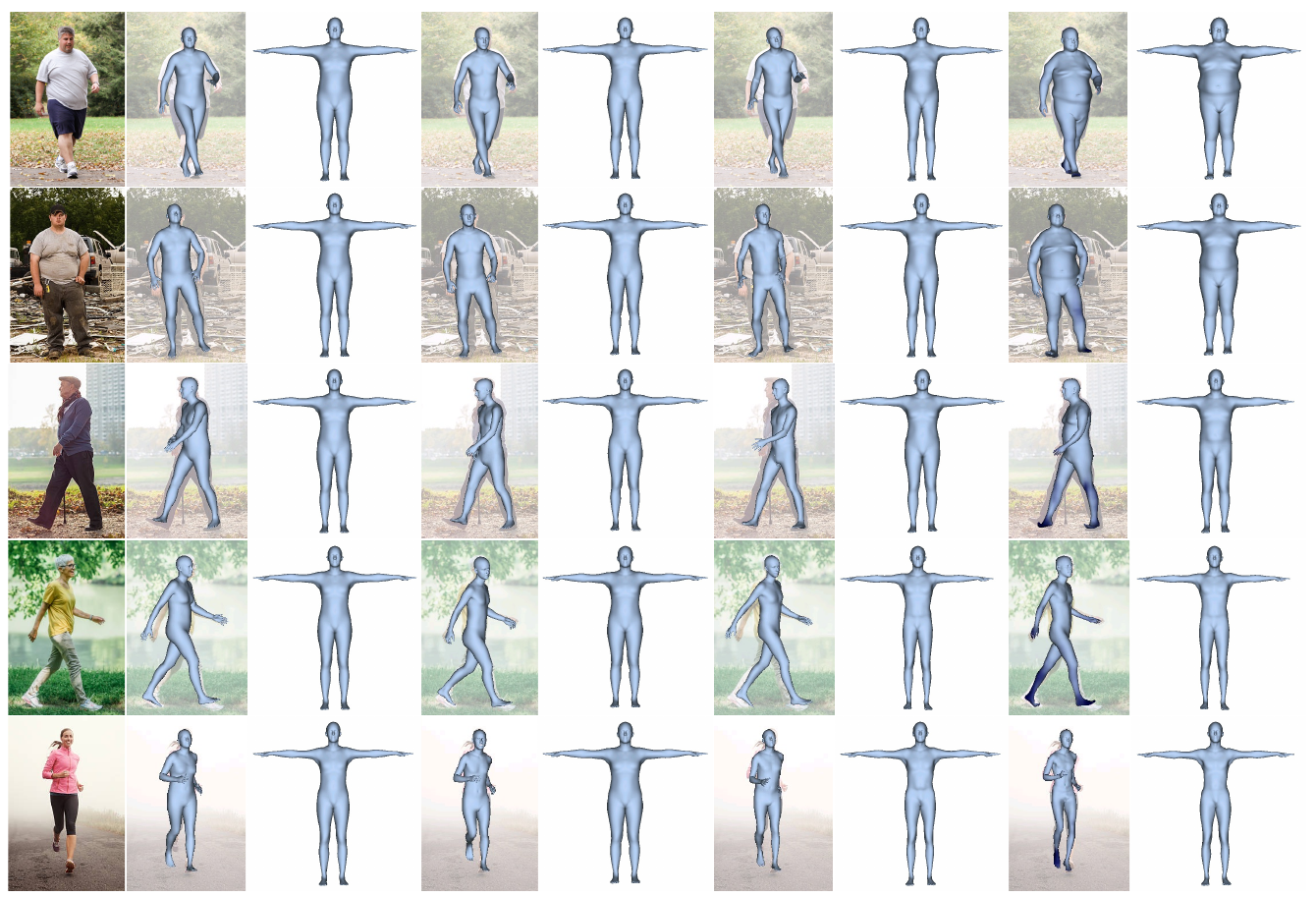}
\put(-340, -9){\footnotesize{Input}}
\put(-308, -9){\footnotesize{FrankMocap~\cite{frankmocap}}}
\put(-220, -9){\footnotesize{ExPose~\cite{expose}}}
\put(-142, -9){\footnotesize{ProHMR~\cite{prohmr}}}
\put(-55, -9){\footnotesize{LVD}}
\vspace{-1mm}
\caption{\footnotesize{{\bf SMPL reconstruction on images on-the-wild.} 
For each method, we show the reconstruction in posed and canonical space. While previous works focus on pose estimation, they are prone to generate always an average body shape. In contrast, LVD generates a much richer distribution of body shapes as shown in the right-most column. 
}
\label{fig:qualitative_results_image}
}
\vspace{-5mm}
\end{figure}
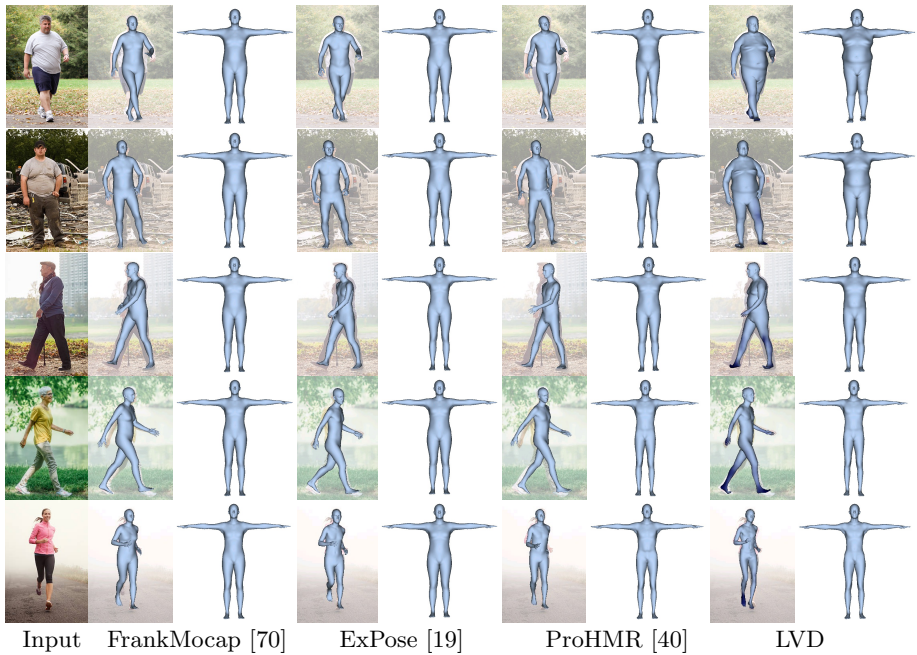

%% file: results_buff_quantitative.tex
\begin{table*}
\vspace{-2mm}
\caption{\small{{\bf Single-view SMPL estimation from LVD and baselines~\cite{smplx,spin,frankmocap,expose,prohmr,sengupta2021hierarchical} in the BUFF Dataset~\cite{buff_dataset}}. The experiments take into account front, side and back views from the original scans and show that LVD outperforms all baselines in all scenarios and metrics except for back views. *We also report the results of PIFu, although note that this is a model-free approach in contrast to ours and the rest of the baselines, which recover the SMPL model. All errors are reported in mm.}
\vspace{-3mm}
\label{tab:quantitative_buff}}
\centering
\resizebox{0.95 \textwidth}{!}{%
\begin{tabular}{l c c c c c c c c c c}
\cmidrule(lr){2-5} \cmidrule(lr){6-9} \cmidrule(lr){10-11}
& \multicolumn{4}{c}{Vertex-to-Vertex} & \multicolumn{4}{c}{Vertex-to-Surface} & V2V & V2S\\
Viewing angle: & 0$^{\circ}$ & 90$^{\circ}$ & 180$^{\circ}$ & 270$^{\circ}$ & 0$^{\circ}$ & 90$^{\circ}$ & 180$^{\circ}$ & 270$^{\circ}$ & Avg. & Avg. \\
\cmidrule(lr){1-1} \cmidrule(lr){2-5} \cmidrule(lr){6-9} \cmidrule(lr){10-11}
*PIFu~\cite{saito2019pifu}  & 36.71 & 40.55 & 72.57 & 39.23 & 35.16 & 39.04 & 71.38 & 38.51 & 47.18 & 46.05 \\
SMPL-X~\cite{smplx} & 41.30	& 77.03	&61.40	& 92.50 & 40.00 & 75.68	& 60.27 &	91.30 & 68.07 & 66.81 \\
SPIN ~\cite{spin}  & 31.96 & 42.10 &53.93 & 44.54 & 30.68 & 40.87 & 52.86 & 43.30 & 43.13 & 41.92 \\
FrankMocap~\cite{frankmocap} & 27.24 & 43.33 & 47.36 & 42.36 & 25.70 & 41.93 & 46.15 & 40.85 & 40.07 & 38.66 \\
ExPose~\cite{expose} & 26.07 & 40.83 & 54.42 & 44.34 & 24.61 & 39.60 & 53.23 & 43.16 & 41.41 & 40.15 \\
ProHMR~\cite{prohmr}  & 39.55 &49.26 & 55.42 & 46.03 & 38.42 & 48.18 & 54.41	& 44.88  & 47.56 & 46.47  \\
Sengupta~\cite{sengupta2021hierarchical}  & 27.70 & 51.10 & {\bf 40.11} & 53.28 & 25.96 &	49.77 & {\bf 38.80} &  52.03 & 43.05 & 41.64 \\
LVD  & {\bf 25.44} & {\bf 38.24} & 54.55 & {\bf 38.10} & {\bf 23.94} & {\bf 37.05} & 53.55 & {\bf 36.94} & {\bf 39.08} & {\bf 37.87} \\
\cmidrule(lr){1-1} \cmidrule(lr){2-5} \cmidrule(lr){6-9} \cmidrule(lr){10-11}
\end{tabular}}
\vspace{-4mm}
\end{table*}

%% file: section_experiments.tex
\vspace{-1mm}
\section{Experiments}\label{sec:exp}
We next evaluate the performance of \Modelname~in the tasks of 3D human reconstruction from a single image and 3D registration. Additionally, we will provide empirical insights about the convergence of the algorithm and its shape expressiveness compared to parametric models.

\vspace{1mm}
\noindent{\bf Data.}
We use the RenderPeople, AXYZ and Twindom datasets~\cite{renderpeople,axyz,twindom}, which consist of 767 3D scans. We first obtain SMPL registrations and manually annotate the correct fits. Then, we perform an aggressive data augmentation by synthetically changing body pose, shape and rendering several images per mesh from  different views and illuminations. By doing, this we collect a synthetic dataset of $\sim600k$ images which we use for training and validation. Test will be performed on real datasets. Please see Suppl. Mat. for more details about the construction of this dataset.

\vspace{-1mm}
\subsection{3D Body shape estimation from a single image}

We evaluate LVD in the task of body shape estimation and compare it against Sengupta~\etal~\cite{sengupta2021hierarchical}, which uses 2D edges and joints to extract features that are used to predict SMPL parameters. We also compare it against a model that estimates SMPL pose and shape parameters given an input image. We use a pre-trained ResNet-18~\cite{resnet} that is trained on the exact same data as LVD. 
This approach fails to capture the variability of body shapes and does not generalize well to new poses. We attribute this to the limited amount of data (only a few hundred 3D scans), with every image being a training data point, while in LVD every sampled 3D point counts as one training example. Figure~\ref{fig:qualitative_results_shape} shows qualitative results on in-the-wild images. The predictions of LVD also capture the body shape better than those of Sengupta~\etal~\cite{sengupta2021hierarchical} and project better to the silhouette of the input person.

Even though our primary goal is not pose estimation, 
we also compare \Modelname~against several recent state-of-the-art  model-based methods~\cite{smplx,spin,frankmocap,expose,prohmr} on the BUFF dataset, which has 9612 textured scans of clothed people. We have uniformly sampled 480 scans and rendered images at four camera views. Table~\ref{tab:quantitative_buff} summarizes the results in terms of the Vertex-to-Vertex (V2V) and Vertex-to-Surface (V2S) distances. The table also reports the results of PIFu~\cite{saito2019pifu}, although we should take this merely as a reference, as this is a model-free approach, while the rest of the methods in the Table are model-based.  Figure~\ref{fig:qualitative_results_image} shows qualitative results on in-the-wild images. With this experiment, we want to show that previous works on pose and shape estimation tend to predict average body shapes. In contrast, our approach is able to reconstruct high-fidelity bodies for different morphotypes. It should be noted that our primary goal is to estimate accurate body shape, and our training data does not include extreme poses. Generalizing LVD to complex poses will most likely require self-supervised frameworks with in the wild 2D images like current SOTA~\cite{hmr,prohmr,expose,spin} 
, but this is out of the scope of this paper, and leave it for future work.

\input{figure_convergence}

Finally, it is worth to point that some of the baselines~\cite{smplx,frankmocap,spin,expose} require 2D keypoint predictions, for which we use the publicly available code of OpenPose~\cite{openpose}. In contrast, LVD requires coarse image segmentations to mask the background out because we trained LVD on renders of 3D scans without background. In any event, we noticed that our model is not particularly sensitive to the quality of input masks, and can still generate  plausible body shapes with noisy masks (see Supp. Mat.).

\input{figure_3d_registration}

\vspace{-1mm}
\subsection{Shape expressiveness and convergence analysis}

We further study the ability of all methods to represent different body shapes. For this, we obtain the SMPL shape for our model and pose estimation baselines in Tab.~\ref{tab:quantitative_buff} and fit the SMPL model with 300 shape components. We then calculate the standard deviation $\sigma_2$ of the second PCA component, responsible for the shape diversity. Fig.~\ref{fig:qualitative_convergence} (left) depicts  the graph of shape $\sigma_2$ vs. V2S error. It is clearly shown that \Modelname~stands out in its capacity to represent different shapes. 
In contrast, most previous approaches have a much lower capacity to recover different body shapes, with a $\sigma_2$ value 3 times smaller than ours.

We also perform an empirical convergence analysis of \Modelname.  Fig.~\ref{fig:qualitative_convergence} (right) plots the average  V2V error (in mm) vs time, computed when performing   shape inference for 200 different samples of the BUFF dataset.   Note that the optimization converges at a tenth of a second using a GTX 1080 Ti GPU. The total computation time is equivalent to 6 iterations of our algorithm. The color-coded 3D mesh on the side of Fig.~\ref{fig:qualitative_convergence} (right) shows in which parts of the body the algorithm suffers the most. These areas are concentrated on the  arms. Other regions that hardly become occluded, such as \textcolor{purple}{torso or head} have the lowest error. The average vertex error is represented with a thicker black line.

\input{results_registration}

Finally, we measure the sensitiveness of the convergence to different initializations of the body mesh. 
We randomly sampled 1K different initializations from the AMASS dataset~\cite{amass} and analized the deviation of the converged reconstructions, obtaining a standard deviation of the SMPL surface vertices of only $\sigma = 1.2$mm across all reconstructions. We credit this robustness to the dense supervision during training, which takes input points from a volume on the 3D space, as well as around the groundtruth body surface. See the Supp. Video for qualitative evidence.

\vspace{-1mm}
\subsection{3D Body Registration}
LVD is designed to be general and directly applicable for different tasks.
We analyze the performance of \Modelname~on the task of  SMPL and SMPL+D registration on  3D point-clouds of humans. This task consists in initially estimating the SMPL mesh (which we do iterating our approach) and then running a second minimization of the Chamfer distance to fit SMPL and SMPL+D. The results are reported in Tab.~\ref{tab:quantitative_scans}, where we compare against LoopReg~\cite{loopreg},  IP-Net~\cite{ipnet}, and also against the simple baseline of registering SMPL with no correspondences starting from a T-Posed SMPL. Besides the V2V and V2S metrics (bi-directional), we also report the Joint error (predicted using SMPL's joint regressor), and the distance between ground truth SMPL vertices and their correspondences in the registered mesh (Vertex distance). Note that again, \Modelname~consistently outperforms the rest of the baselines. This is also qualitatively shown in Fig.~\ref{fig:qualitative_results_scans}.%

\input{figure_hand_registration}

\subsection{3D Hand Registration}
The proposed approach is directly applicable to any statistical model, thus we also test it in the task of registration of MANO~\cite{mano} from input  point-clouds of hands, some of them incomplete. For this experiment, we do not change the network hyperparameters and only update the number of vertices to predict (778 for MANO). We test this task on the MANO~\cite{mano} dataset, where the approach also outperforms IP-Net\cite{ipnet}, trained on the same data. Tab.~\ref{tab:quantitative_scans_mano} summarizes the performance of LVD and baselines, and qualitative examples are shown in Fig.~\ref{fig:fitting_hands}. Note that LVD shows robustness even in situations with partial point clouds.

%% file: figure_convergence.tex
\begin{figure}[t]
\vspace{-3mm}
\centering
\resizebox{0.4\textwidth}{0.3\textwidth}{
\begin{tikzpicture}
\begin{axis}[
    xlabel={Shape $\sigma$},
    ylabel={V2S Reconstruction error},
    xmin=0.05, xmax=1.4,
    ymin=32, ymax=73,
    legend pos=north west,
    ymajorgrids=true,
    grid style=dashed,
    legend style={font=\small},
]
\tikzstyle{every node}=[font=\small]
\node[label={SMPLify},circle,fill,color=pink,inner sep=2pt] at (axis cs:0.87,66.82) {};
\node[label={[xshift=0.0cm, yshift=-0.04cm] SPIN},circle,color=purple,fill,inner sep=2pt] at (axis cs:0.378, 41.92) {};
\node[label={[xshift=0.3cm, yshift=-0.65cm]FrankMocap},circle,color=black,fill,inner sep=2pt] at (axis cs: 0.258, 38.66) {};
\node[label={[xshift=0.8cm, yshift=-0.5cm]ExPose},circle, color=red, fill, inner sep=2pt] at (axis cs: 0.4515,40.15) {};
\node[label={ProHMR},circle,fill, color=green, inner sep=2pt] at (axis cs: 0.3565,46.47) {};
\node[label={[xshift=0.2cm, yshift=0cm]Sengupta},circle, color=olive, fill, inner sep=2pt] at (axis cs: 0.713,41.64) {};
\node[label={[xshift=0.0cm, yshift=-0.65cm] LVD},circle,fill,color=blue,inner sep=2pt] at (axis cs:1.279, 37.87) {};
\end{axis}
\end{tikzpicture}}
\hspace{1cm}
\includegraphics[width=0.48\linewidth, , trim={0cm, 0cm 0.3cm 0cm}, clip =true]{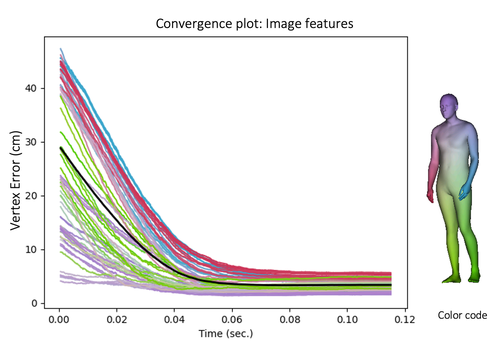}\\
\vspace{-3mm}
\caption{\small{
\textit{Left:} Variability of predicted body shape parameters (x-axis) with respect to vertex error (y-axis, lower is better) for works that fit SMPL to images. Previous approaches have mostly focused on the task of pose estimation. LVD, instead, aims to represent a more realistic distribution of predicted body shapes.
\textit{Right:} Convergence analysis of the proposed optimization, showing the distance from each SMPL vertex to the groundtruth scan during optimization, averaged for 200 examples of the BUFF dataset. The first iteration also includes the time to obtain the local representations used during the rest of the optimization. 
Each line color encodes a different body region and the black line shows the average error of all vertices. %
\label{fig:qualitative_convergence}
}}
\vspace{-3mm}
\end{figure}

%% file: figure_3d_registration.tex
\begin{figure}[t]
\centering
\includegraphics[width=1.0\linewidth, trim={0.05cm, 0.05cm 0.1cm 0.05cm}, clip =true]{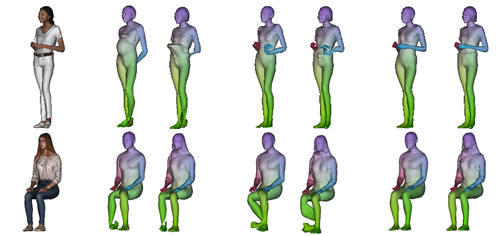}
\put(-340, -9){\small{Input scan}}
\put(-265, -9){\small{LoopReg~\cite{loopreg}}}
\put(-164, -9){\small{IP-Net~\cite{ipnet}}}
\put(-53, -9){\small{Ours}}
\vspace{-1mm}
\caption{\small{SMPL and SMPL+D registration of 3D scans from LVD in comparison to LoopReg and IP-Net.%
}}
\label{fig:qualitative_results_scans}
\end{figure}
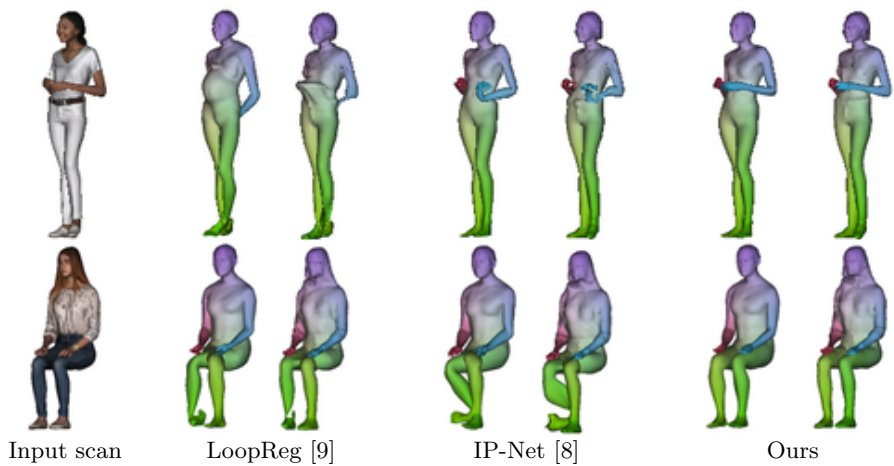

%% file: results_registration.tex
\begin{table}[t]
\caption{\small{{\bf Evaluation on SMPL and SMPL+D registration on the RenderPeople Dataset~\cite{renderpeople}}. 
The initial SMPL estimation from LVD is already very competitive against baselines~\cite{loopreg,ipnet}.
By using these predictions as initialization for SMPL/SMPL+D registration, we obtain $\sim28.4\%$ and $\sim37.7\%$ relative improvements with respect to the second-best method\cite{ipnet} in joint and SMPL vertex distances respectively.
}
\label{tab:quantitative_scans}}
\vspace{-1mm}
\centering
\resizebox{0.95 \textwidth}{!}{%
\begin{tabular}{r r c c c c c c c c c}
\cmidrule(lr){3-11}
& & Forward pass & \multicolumn{4}{c}{SMPL Registration} &  \multicolumn{4}{c}{SMPL+D Registration} \\
\cmidrule(lr){3-3} \cmidrule(lr){4-7} \cmidrule(lr){8-11}
& & LVD & No corresp. & LoopReg~\cite{loopreg} & IP-Net~\cite{ipnet} & \multicolumn{1}{c}{LVD} & No corresp. & LoopReg~\cite{loopreg} & IP-Net~\cite{ipnet} & \multicolumn{1}{c}{LVD} \\
\cmidrule(lr){3-3} \cmidrule(lr){4-7} \cmidrule(lr){8-11}
\multirow{2}{*}{\begin{tabular}[r]{@{}c@{}}SMPL \\error \end{tabular}} & Joint [cm] & \multicolumn{1}{c}{5.89} & 16.6 & 9.33 & 3.60 & \multicolumn{1}{c}{\bf 2.53} & 16.6 & 9.32 & 3.63 & {\bf 2.60} \\
& Vertex [cm] & \multicolumn{1}{c}{6.27} & 21.3 & 12.2 & 5.03 & \multicolumn{1}{c}{\bf 3.00} & 21.3  & 12.3 & 5.20 & {\bf 3.24} \\
\cmidrule(lr){1-11}
\multirow{2}{*}{\begin{tabular}[c]{@{}l@{}}Recons.\\ to Scan \end{tabular}} & V2V [mm] & \multicolumn{1}{c}{8.98} & 12.51 & 10.35 & 8.84 & \multicolumn{1}{c}{\bf 8.16} &  1.45& 1.43  & 1.21 & {\bf 1.14 }\\
& V2S [mm] & \multicolumn{1}{c}{6.61} & 10.53 & 8.19 & 6.61 & \multicolumn{1}{c}{\bf 5.87} & 0.72& 0.69& 0.53 & {\bf 0.47 }\\
\cmidrule(lr){1-11}
\multirow{2}{*}{\begin{tabular}[c]{@{}l@{}}Scan to\\ Recons.\end{tabular}} & V2V [mm] & \multicolumn{1}{c}{12.6} & 16.92 & 14.27 & 12.25 & \multicolumn{1}{c}{\bf 11.31} &8.53 & 8.01& 7.22& {\bf 6.88 }\\
& V2S [mm] & \multicolumn{1}{c}{9.31} & 13.75 & 10.49 & 8.45 & \multicolumn{1}{c}{\bf 7.43} &4.22 & 3.47& 2.78& {\bf 2.44 }\\
\cmidrule(lr){1-11}
\end{tabular}}
\vspace{-4mm}
\end{table}

%% file: figure_hand_registration.tex
\begin{figure}[t]
\vspace{-1mm}
\centering
\includegraphics[width=1.0\linewidth, trim={0.05cm, 0.05cm 0.1cm 0.05cm}, clip =true]{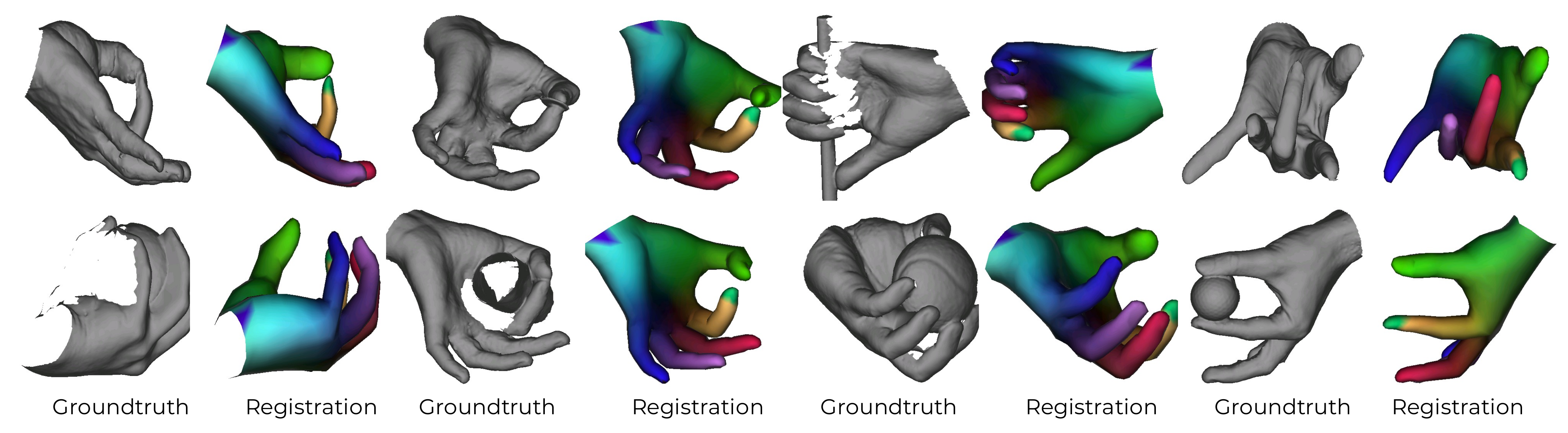}
\vspace{-1mm}
\caption{\small{Registration of MANO from input pointclouds. We include more visuals and qualitative comparisons with baselines in Supplementary Material.
}
\label{fig:fitting_hands}
}
\vspace{-1mm}
\end{figure}

\begin{table}[t]
\caption{\small{Registration of MANO~\cite{mano} from input 3D pointclouds of hands.}}
\label{tab:quantitative_scans_mano}
\vspace{1mm}
\centering
\resizebox{0.95 \textwidth}{!}{%
\begin{tabular}{l c c c c c c }
\cmidrule(lr){2-7}
& \multicolumn{2}{c}{MANO Error} & \multicolumn{2}{c}{Reconstruction to Scan} &  \multicolumn{2}{c}{Scan to Reconstruction} \\
\cmidrule(lr){2-3} \cmidrule(lr){4-5} \cmidrule(lr){6-7}
Method & Joint [cm]  & Vertex [cm] & V2V [mm]  & V2S [mm] & V2V [mm] & V2S [mm] \\
No corresp. & 6.49 & 7.05 &  5.31 &  5.28 & 8.06 & 6.40 \\
IP-Net~\cite{ipnet} & 1.44 & 1.73 & 3.29 & 3.23 & 6.17 & 4.08 \\
LVD & {\bf .76} & {\bf .96} & {\bf 2.73} & {\bf 2.65} & {\bf 5.62} &  {\bf 3.33} \\

\cmidrule(lr){1-1} \cmidrule(lr){2-7}
\end{tabular}}
\vspace{-2mm}
\end{table}

%% file: section_conclusion.tex
\section{Conclusion}

We have introduced Learned Vertex Descent, a novel framework for human shape recovery where vertices are iteratively displaced towards the predicted body surface. The proposed method is lightweight, can work real-time and surpasses previous state-of-the-art in the tasks of body shape estimation from a single view or 3D scan registration, of both the full body and hands. Being so simple, easy to use and effective, we believe LVD can be an important building block for future model-fitting methods. 
Future work will focus in self-supervised training formulations of LVD for predicting body shape in difficult poses and scenes, and tackling multi-person scenes efficiently.

%% file: supplementary.tex
\pagestyle{headings}

\title{SUPPLEMENTARY MATERIAL}
\author{Enric Corona$^{1}$ \and \hspace{0.5mm}
Gerard Pons-Moll$^{2,3}$ \and \\ \hspace{0.5mm} Guillem Aleny\`a$^{1}$ \and Francesc Moreno-Noguer$^{1}$}
\institute{\hspace{-4mm}${}^{1}$Institut de Robòtica i Informàtica Industrial, CSIC-UPC, Barcelona, Spain\\${}^{2}$University of T\"{u}bingen, Germany, ${}^{3}$Max Planck Institute for Informatics, Germany}

\titlerunning{Learned Vertex Descent: A New Direction for 3D Human Model Fitting}
\authorrunning{Corona et al.}
\maketitle

In this supplementary material, we provide a detailed description of the implementation details and the data augmentation we used. We also include more qualitative examples and a supplementary video which summarizes the method and the contributions of the paper.

\section{Implementation details}\label{sec:implem}
We next describe the main implementation details. 
The code will be made publicly available.

The clipping factor for the learnt gradient is to $18\%$  of the vertical size of the scan, which we normalize between $-0.75$ and $0.75$.
In our experiments, $H=W=256$, $f$ is a stacked hourglass network~\cite{newell2016stacked} trained from scratch with 4 stacks and batch normalization replaced with group normalization~\cite{wu2018group}. 
The feature embeddings have size $128\times128$ with 256 channels each. Therefore, query points have a feature size of $F = 256\times{4} = 1024$.
The MLP $f$ is formed by 3 fully connected layers with Weight Normalization~\cite{salimans2016weight}, and deeper architectures or positional encoding did not help to improve performance. We attribute this to the fact that the MLP is already obtaining very rich representations from feature maps. For images, we assume a weak-perspective projection although our approach is compatible with perspective cameras.

The networks are trained end-to-end with batch size $4$, learning rate 0.001 during 500 epochs, and then with linear learning rate decay during 500 epochs more. We use Adam Optimizer~\cite{adam} with $\beta_1=0.9$ $\beta_2=0.999$. 
When considering point-clouds as input we train an IF-Net backbone\cite{ifnet} from scratch with the same training conditions and number of iterations.

Implementation-wise $f$ has an output dimension of $N=6890$. When estimating an SMPL shape, we input a surface of $6890\times3$ and obtain a prediction tensor with shape $6890\times6890\times3$, from which we sample the diagonal to obtain per-vertex displacements ($6890\times3$) and move each vertex in the correct direction.
For the task of registration of the MANO model~\cite{mano}, we instead predict 778 vertices.

To compare LVD against other baselines, we used their available code. For SMPL-X, we fitted the SMPL model for better comparison with ours and previous works, using their most recent code (SMPLify-X) with the variational prior.

\input{figure_convergence_supp}

\begin{figure*}
\noindent\includegraphics[width=1\linewidth, trim={0.1cm, 0.1cm 0.1cm 0.1cm}, clip =true]{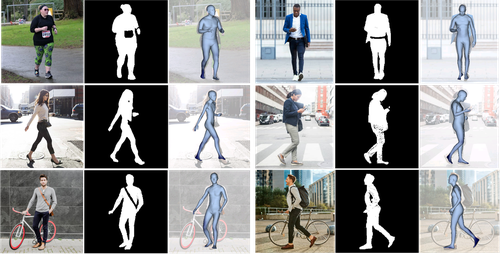}\\
\put(16, 0){\small{Image}}
\put(75, 0){\small{Mask}}
\put(138, 0){\small{LVD}}
\put(194, 0){\small{Image}}
\put(254, 0){\small{Mask}}
\put(310, 0){\small{LVD}}
\caption{{\bf SMPL reconstruction on images in-the-wild,} and the predicted foreground masks\cite{rp-r-cnn}. Even with noisy segmentations, the predicted SMPL accurately represents the body shapes and poses of the target people.
}
\label{fig:masks}
\end{figure*}

\input{figure_image_reconstruction_supp}
\input{figure_3d_registration_supp}

\input{figure_hand_reconstruction_supp}

\input{figure_failure_cases_supp}

\section{Data Augmentation for image data}

As mentioned in the main document, we use the RenderPeople, AXYZ and Twindom datasets~\cite{renderpeople,axyz,twindom}, which consist of 767 3D scans. We first obtain SMPL registrations and manually annotate the correct fits, leaving 750 scans. 
Due to  the reduced number of 3D scans, we augment each of them by changing its pose and body shape. On one side, we label pose vectors for humans walking and running, and automatically select a random pose + noise for each new augmentation. To pose the 3D scan, we simply assign the skinning weights of each 3D surface vertex to those of the closest SMPL vertex. This can lead to several artifacts, for body parts that are in contact, such as hands, which will generate very large triangles. We manually prune the generated 3D scans to remove these cases.

Next, we tune the body shape of each 3D scan by changing the first shape parameter in the PCA space. We discretize a number of augmentations with respect to the initial shape and calculate the linear displacement for each body vertex. For the 3D scan we apply the displacement of the closest vertex. This augmentation is proven to be really useful and does not significantly create artifacts since it retains self-contact information. We perform 6 augmentations for each scan.

For the task of human reconstruction from images, we then render each augmentation by rotating around the yaw axis to gather views with different illuminations. As mentioned in the main document, in total we obtain $\sim 680k$ rendered images that are used for training and validation. 

Note that the original data consisted only of a few hundred  3D scans, all with very average body shapes. The augmentation led the model represent more diverse shapes and avoid overfitting, but the proposed Learned Vertex Descent paradigm was necessary for it to represent them well. The baseline that predicts SMPL parameters directly did not manage to generalize well beyond the training set.

\section{Experiments}

As mentioned in the main document, we train our model without backgrounds when taking images as input. Therefore at test time we  use RP-R-CNN~\cite{rp-r-cnn} to automatically segment the foreground person before running the forward pass.
However, this can still generate masks with artifacts or missing parts. We show in Fig.~\ref{fig:masks} that the proposed approach is robust to these noisy masks or parts that were incorrectly segmented.

We also show more qualitative examples of 3D reconstruction from a single view in-the-wild in Fig.~\ref{fig:qualitative_results_image_supp}, and Fig.~\ref{fig:more_comparisons_supp} shows comparisons with the rest of the methods that are not shown in the main document. In particular, we noted several differences between optimization-based and learning-based body pose/shape estimation methods. On one hand, optimization-based methods~\cite{bogo2016keep,smplx} are often accurate, but have severe failure cases and are slow. On the other hand, learning based methods~\cite{spin,frankmocap,expose,prohmr} regress global parameters from the full image. Hence, the shape estimates have a strong bias towards the mean. Moreover, 
learning-based methods are not able to verify their initial estimates against the image.

Our goal in this paper is to combine the advantages of both methods. LVD produces varied shape estimates thanks to the learned per vertex descent directions which are conditioned on local image evidence, and can work in real time.

In addition, we focus on designing a general method that is straightforward to apply to other input modalities such as 3D point clouds. In this direction,
Fig.~\ref{fig:qualitative_results_scans_supp} includes more results on the task of 3D registration of 3D scans and Fig.~\ref{fig:qualitative_results_hand_supp} shows 3D registration results of MANO of LVD in comparison to those of IP-Net~\cite{ipnet}. IP-Net obtains quantitative results close to LVD, and works generally well for clean 3D scans. However, it might converge to wrong local minima when tackling 3D point clouds with objects or holes.

\section{Failure cases.} \label{sec:failure}
We finally include failure cases of LVD in all tasks where we evaluate our approach, in Fig.~\ref{fig:failure_cases_supp}. For the task of body shape estimation from single view (First row), the body shapes we can generate are limited by the SMPL model and the training data, and cannot accurately reproduce body shapes of \eg pregnant women (second example). Furthermore, our training data is rather limited in the diversity of body poses, so challenging body poses is another reason for failure cases. For instance, examples in Fig.~\ref{fig:failure_cases_supp} top-left and top-right show scenarios that are rare in the train data, and the predicted body does not correctly adjust to the input image. However, note that the wrong body parts are predicted to have a big uncertainty (in dark blue).

In Fig.~\ref{fig:failure_cases_supp} (Second row) we show more examples of failure cases in 3D registration of human scans and hands.

\clearpage

%% file: figure_convergence_supp.tex
\begin{figure*}
\noindent\includegraphics[width=1.0\linewidth, , trim={0cm, 0cm 0.3cm 0cm}, clip =true]{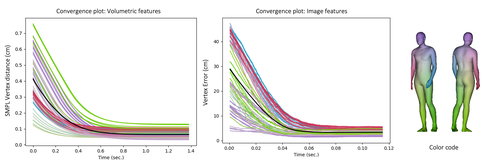}
\caption{{\bf Convergence plot of the proposed optimization, for voxel-based experiments in comparison to image-based reconstruction}. 
In comparison with the reported results on image-based reconstruction (which also are shown in the main paper), volumetric reconstruction takes almost a second to converge with our settings.
Experiments were run on a single GeForce\textsuperscript{\textregistered} GTX 1080 Ti GPU. The black line represents the average of all vertex errors while the remaining colors show how the error is distributed among different body parts, \eg. arms and feet accumulate the biggest error while torso or head generally are the most accurately reconstructed parts.
}
\label{fig:qualitative_convergence_supp}
\end{figure*}

%% file: figure_image_reconstruction_supp.tex
\begin{figure*}[t]
\noindent\includegraphics[width=1\linewidth, trim={0.1cm, 0.1cm 0.1cm 0.1cm}, clip =true]{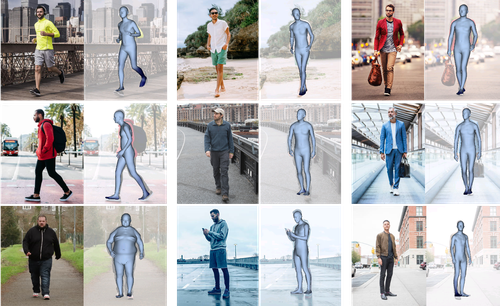}
\caption{{\bf More examples of body shapes estimated on images in-the-wild.}
\label{fig:qualitative_results_image_supp}
}
\end{figure*}

\begin{figure}[t]
\noindent\includegraphics[width=1\linewidth, trim={0.1cm, 0.1cm 0.1cm 0.1cm}, clip =true]{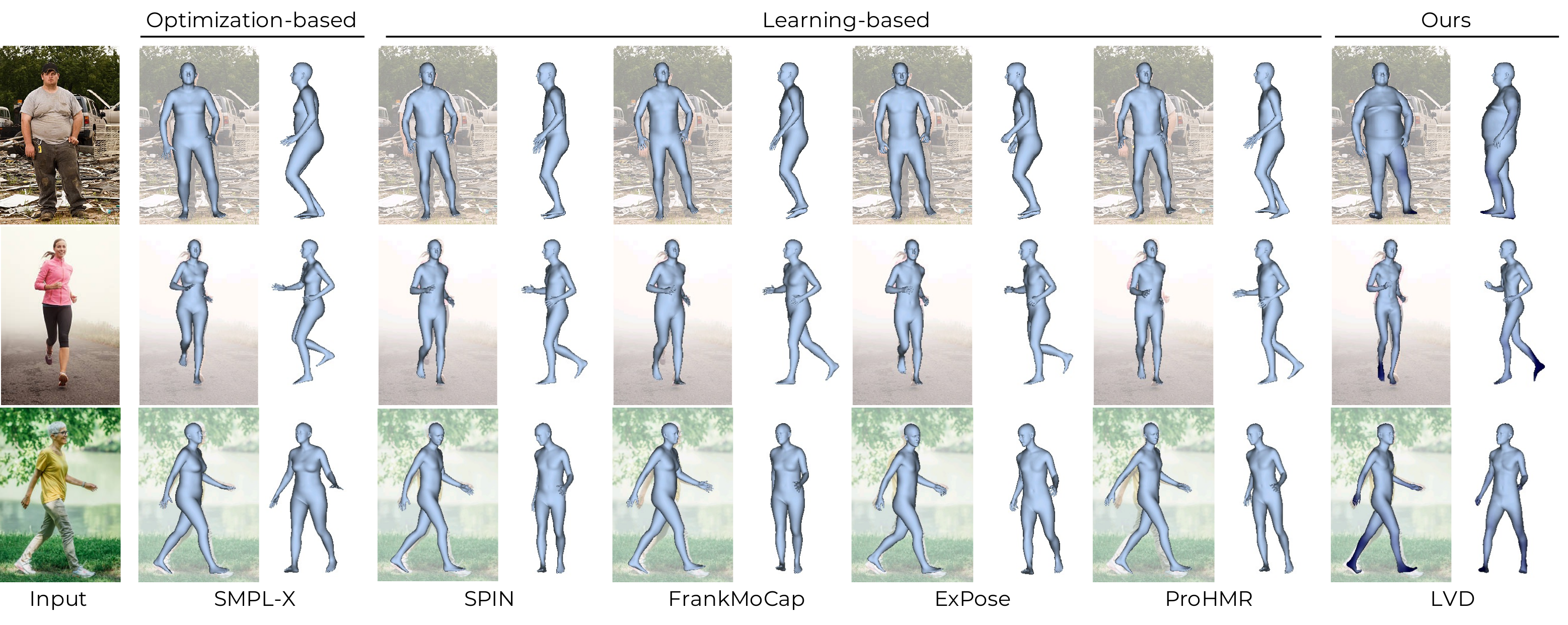}
\caption{{\bf Qualitative comparisons with more methods.} For each method, we show front and side views of the reconstruction.
\label{fig:more_comparisons_supp}
}
\end{figure}

%% file: figure_3d_registration_supp.tex
\begin{figure*}
\noindent\includegraphics[width=0.95\linewidth, trim={0.05cm, 0.05cm 0.1cm 0.1cm}, clip =true]{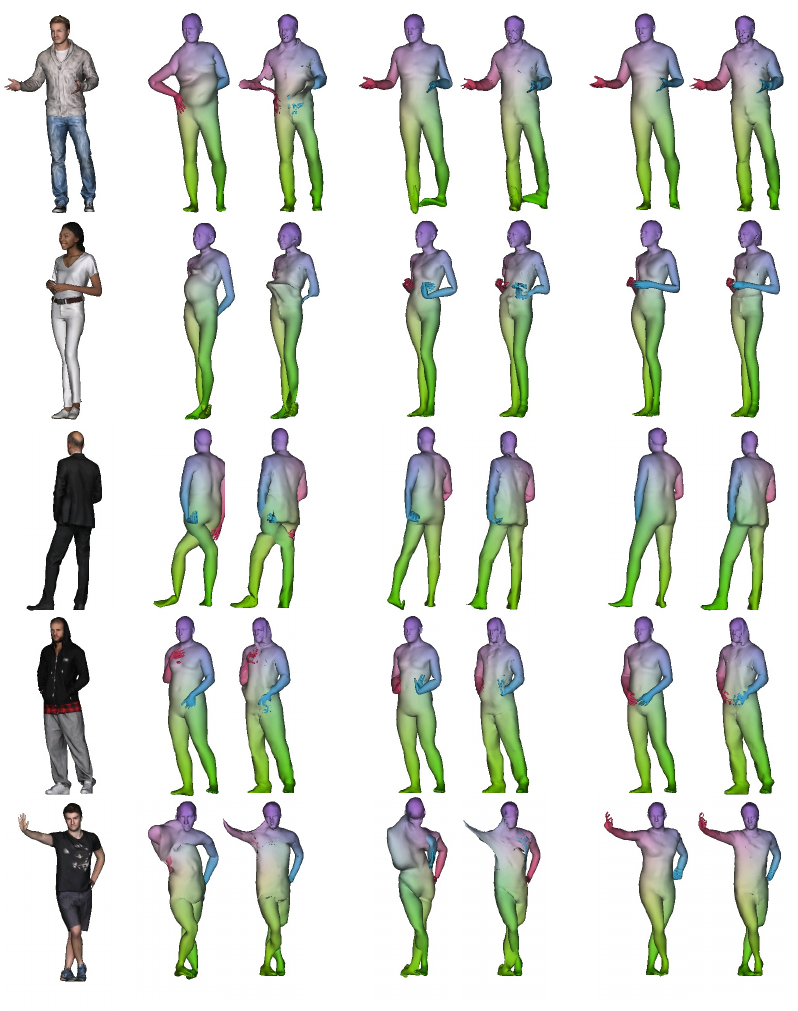}\\
\put(21, 0){\small{Scan}}
\put(74, 0){\small{LoopReg~\cite{loopreg}}}
\put(174, 0){\small{IP-Net~\cite{ipnet}}}
\put(283, 0){\small{Ours}}
\caption{SMPL registration of 3D scans showing SMPL and SMPL-D for LoopReg, IP-Net and LVD. 
}
\label{fig:qualitative_results_scans_supp}
\end{figure*}

%% file: figure_hand_reconstruction_supp.tex
\begin{figure*}
\noindent\includegraphics[width=1\linewidth, trim={0.1cm, 0.1cm 0.1cm 0.1cm}, clip =true]{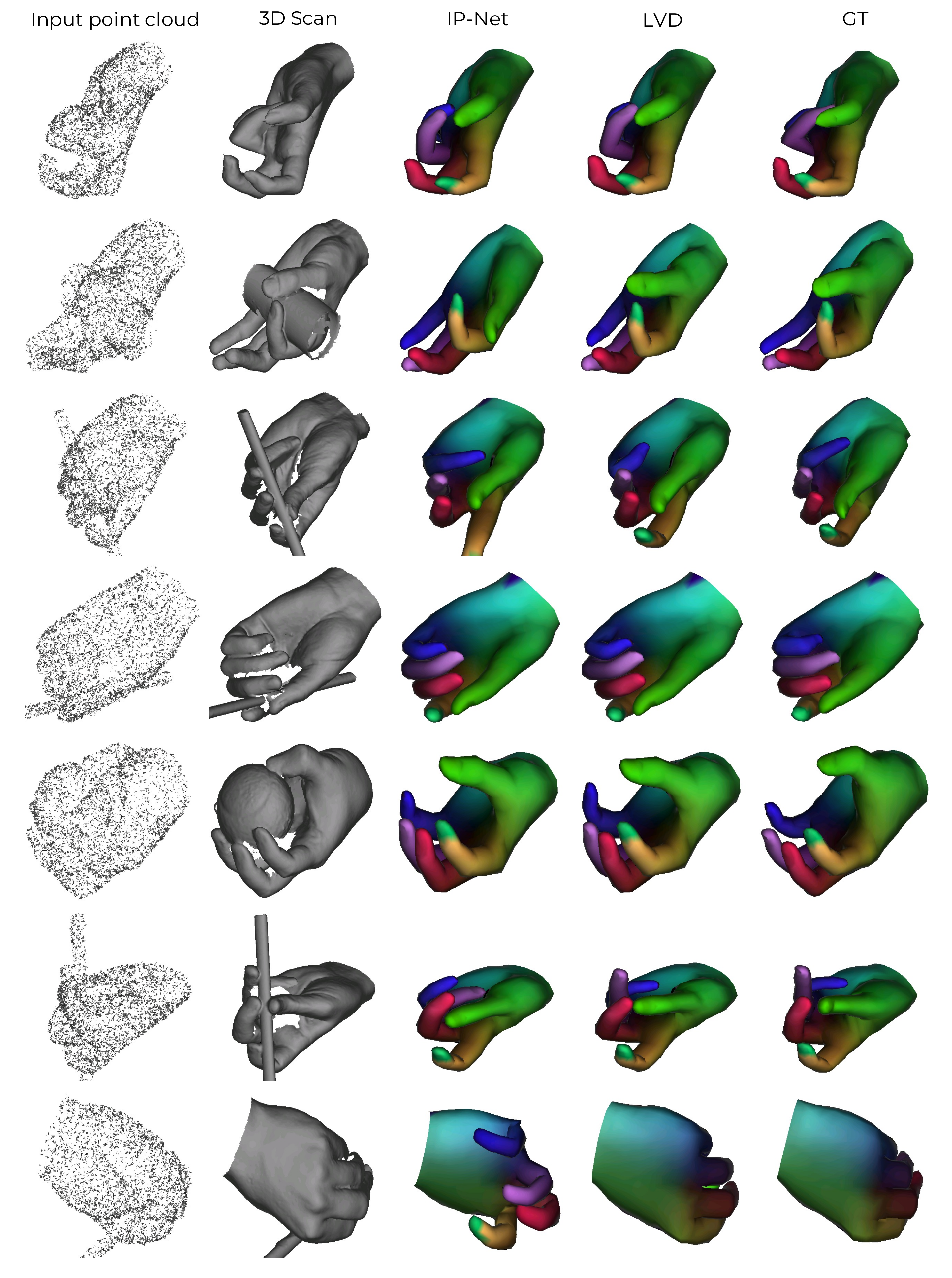}
\caption{{\bf Registration of 3D Hands using MANO~\cite{mano}.} The input to IP-Net~\cite{ipnet} or LVD is the input point cloud in the left column, while the groundtruth 3D scan is shown in the second column. IP-Net performs similarly well in most cases, but is most confused in the presence of other objects or very noisy pointclouds.
\label{fig:qualitative_results_hand_supp}
}
\end{figure*}

%% file: figure_failure_cases_supp.tex
\begin{figure}[t]
\noindent\includegraphics[width=1\linewidth, trim={0.1cm, 0.1cm 0.1cm 0.1cm}, clip =true]{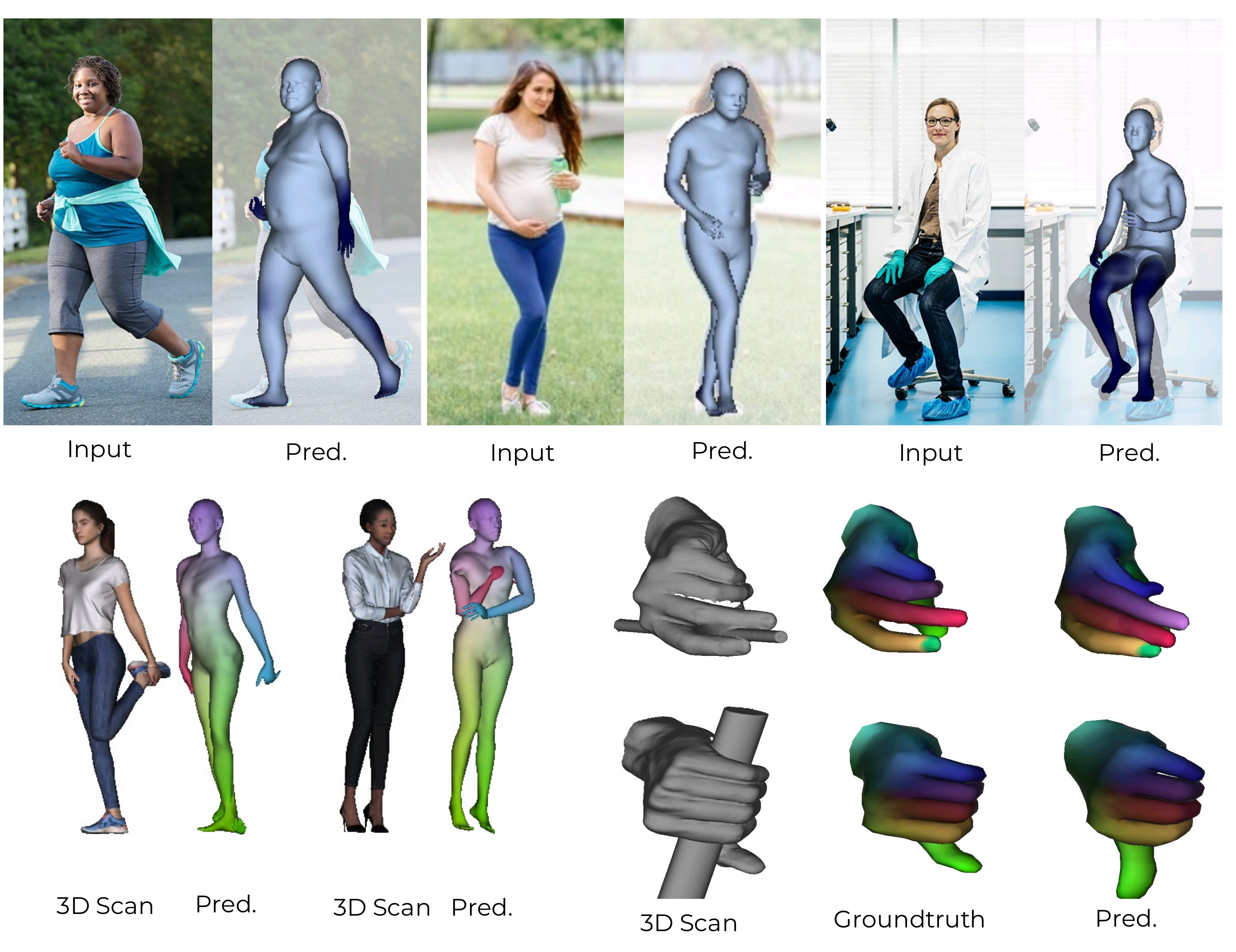}
\caption{{\bf Failure cases from LVD in body shape estimation from single view images (first row), 3D registration of humans from point clouds (second row - left) and 3D registration from hands (second row - right).} See Section~\ref{sec:failure} for more details.
\label{fig:failure_cases_supp}
}
\end{figure}